\documentclass[11pt]{article}
\usepackage{acl}
\usepackage{placeins}

\usepackage[english,bidi=default]{babel}
\babelfont{rm}[
  Extension=.otf,
  UprightFont=*-regular,
  BoldFont=*-bold,
  ItalicFont=*-italic,
  BoldItalicFont=*-bolditalic
]{texgyretermes}
\babelfont{sf}[
  Extension=.otf,
  UprightFont=*-regular,
  BoldFont=*-bold,
  ItalicFont=*-italic,
  BoldItalicFont=*-bolditalic
]{texgyreheros}
\babelfont{tt}[
  Extension=.otf,
  UprightFont=*-Regular,
  BoldFont=*-Bold
]{Inconsolatazi4}

\usepackage{graphicx}
\usepackage{twemojis}
\newcommand{\emoji}[1]{%
  \ifthenelse{\equal{#1}{money-bag}}{\texttwemoji{money bag}}{%
  \ifthenelse{\equal{#1}{drop-of-blood}}{\texttwemoji{drop of blood}}{%
  \ifthenelse{\equal{#1}{gem-stone}}{\texttwemoji{gem stone}}{%
  \ifthenelse{\equal{#1}{chart-increasing}}{\texttwemoji{chart increasing}}{%
  \ifthenelse{\equal{#1}{chart-decreasing}}{\texttwemoji{chart decreasing}}{%
  \ifthenelse{\equal{#1}{check-mark-button}}{\texttwemoji{check mark button}}{%
  \ifthenelse{\equal{#1}{right-arrow}}{\texttwemoji{right arrow}}{%
  \ifthenelse{\equal{#1}{down-arrow}}{\texttwemoji{down arrow}}{%
  \ifthenelse{\equal{#1}{up-arrow}}{\texttwemoji{up arrow}}{%
  \ifthenelse{\equal{#1}{face-with-tears-of-joy}}{\texttwemoji{face with tears of joy}}{%
  \ifthenelse{\equal{#1}{thinking-face}}{\texttwemoji{thinking face}}{%
  \ifthenelse{\equal{#1}{thumbs-up}}{\texttwemoji{thumbs up}}{%
  \texttwemoji{#1}}}}}}}}}}}}}%
}

\usepackage{url}
\usepackage{latexsym}
\usepackage{microtype}

\usepackage{siunitx}
\usepackage{enumitem}
\usepackage{threeparttable}
\usepackage{makecell}
\usepackage{multirow}
\usepackage{booktabs}
\usepackage{adjustbox}
\usepackage{subcaption}

\title{Cross-Cultural Transfer of Emoji Semantics and Sentiment in Financial Social Media}

\author{
Ahmed Mahrous \hfill Roberto Di Pietro \\
King Abdullah University of Science and Technology (KAUST) \\
\texttt{\{ahmed.mahrous, roberto.dipietro\}@kaust.edu.sa}
}

\begin{document}
\maketitle

\begin{abstract}
Emojis are widely used in online financial communication, but it is unclear whether they provide transferable sentiment signals across languages, platforms, and asset communities. This study examines the extent to which emoji usage, semantics, and sentiment polarity remain stable across financial communities, and how these layers influence zero-shot sentiment transfer. Using large corpora of Twitter and StockTwits posts in four languages, we measure cross-community divergence and evaluate sentiment models trained under emoji-only, text-only, and text+emoji inputs.

We find that emoji frequencies differ across communities, especially across languages, but their semantics and sentiment polarity are largely stable. Cross-asset transferability shows minimal degradation, while cross-language transfer remains the most challenging. Including emojis consistently reduces transfer gaps relative to text-only models. These results indicate that financial communication exhibits a partially shared ``emoji code,'' and that emojis provide compact, language-independent sentiment cues that improve model generalization across markets and platforms.
\end{abstract}

\section{Introduction}

Emojis are widely used in digital communication, appearing in more than one in five tweets and billions of messages daily \cite{EmojipediaStats}. Because they compress emotion and meaning, emojis have become important features in NLP tasks such as sentiment analysis. Their visual and symbolic nature also suggests the possibility of language-independent sentiment signals.

However, prior work shows substantial variation in emoji usage and interpretation across cultures, demographics, and platforms \cite{lu2016emoji, ljubesic2016global, miller2016variation, miller2017ambiguity, guntuku2019cultural, bai2019review}. Emojis that seem universal often acquire community-specific meanings, limiting the ability of sentiment models to generalize across domains.

Financial communication offers a focused setting to examine this question. Investor communities form a global subculture with shared jargon and symbolic cues. Emojis such as \emoji{gem}, \emoji{rocket}, and \emoji{drop-of-blood} often convey notions of persistence, optimism, or loss. If these meanings are stable across languages and platforms, sentiment models trained in one financial community may transfer more effectively to others.

Prior studies have shown that emoji-based sentiment models can match or exceed text-only models while being faster, cheaper, and more interpretable---features valuable for real-time trading and high-frequency analytics \cite{colavito2025benchmarking, di2024performance}. Understanding whether emoji semantics remain stable across languages and platforms is, therefore, practically relevant for building generalizable financial NLP systems. 

While earlier work has examined emoji universality in general or non-financial contexts, little is known about how emojis behave within financial communities. Existing evidence shows that emoji usage in finance differs from everyday communication \cite{mahrous2023microblogs}, but it remains unclear whether these domain-specific patterns affect the cross-community transferability of emoji meaning and sentiment.

To address this gap, we study whether emoji-based sentiment representations learned in one financial community generalize to others, and how linguistic, asset-specific, and platform-specific differences influence this transferability. Specifically, we ask:
\begin{enumerate}[label=(\roman*)]
    \item Do financial communities use emojis with comparable frequency, meaning, and affective associations across languages, platforms, and asset classes? 
    \item How do these differences affect the zero-shot transfer of financial sentiment models? 
\end{enumerate}

We provide a large-scale empirical analysis across multiple languages, assets, and platforms, and show that financial communities exhibit both shared and community-specific emoji patterns, with consequences for cross-domain sentiment transfer.

\section{Related Work}

Emojis have been described as a ``ubiquitous language'' transcending linguistic boundaries \cite{lu2016emoji}, yet substantial evidence shows that their usage and interpretation vary across cultures, platforms, and communities. Studies report clear regional differences in emoji frequency and emotional tone \cite{ljubesic2016global, guntuku2019cultural}, as well as culture-specific uses and meanings \cite{sampietro2022kissing, togans2021crosscultural, sun2023cultural}. Basic facial emojis may generalize across languages, but many symbols exhibit localized semantics \cite{barbieri2016cosmopolitan, barbieri2018semeval, tomihira2020bert}. Variation also appears within communities: platform renderings can induce sentiment disagreement, and textual context may shift emoji interpretation \cite{miller2016variation, miller2017ambiguity}. Demographic and individual factors further contribute to heterogeneous sentiment perception \cite{koch2022demographics, chen2024individual, hakami2021arabic}.

In finance, emojis have emerged as meaningful sentiment indicators. Prior work shows that incorporating emojis can improve sentiment classification accuracy and efficiency in financial microblogs \cite{chenxing2023finance, mahrous2023microblogs, mahmoudi2022domainspecific}, and that finance-specific emoji embeddings provide explanatory power for price movements and investor sentiment \cite{reschke2024stockreturns, carrillo2022investors}. These findings indicate that emojis carry substantive information in financial communication. 

While prior work establishes that emojis are informative features for financial sentiment analysis and market prediction, it remains limited in scope along three dimensions. First, most studies focus on a single platform (e.g., Twitter or StockTwits) or a single asset class, without evaluating cross-platform or cross-asset generalization. Second, existing analyses are largely restricted to English or monolingual settings, leaving the cross-lingual behavior of financial emojis underexplored. Third, prior work evaluates in-domain performance, without examining whether emoji-based representations transfer across communities under a distribution shift. In contrast, our study systematically evaluates emoji usage, semantics, and sentiment across languages, platforms, and assets, and measures zero-shot transfer performance, offering a view of the generalizability of emoji models in financial text. It remains unclear whether financial emojis exhibit the same degree of heterogeneity observed in general discourse and how such variation affects the transfer of cross-community sentiment analysis models.

\section{Data} 
\subsection{Sources}
We analyze finance-related microblogs from Twitter and StockTwits, totaling over 100M messages across more than a decade. The Twitter data combines two public datasets \cite{penasmartinez2023kaggle, alaix14kaggle}. For StockTwits, we use one dataset covering equities \cite{divernois2024stocktwits} and an additional cryptocurrency dataset collected by the authors. The corpora span four languages (EN, ES, JA, TR) and two asset categories (stocks and cryptocurrencies). 

The dataset comprises over 100 million messages in total, including approximately 21M English tweets, 0.65M Japanese tweets, 0.25M Spanish tweets, and 0.23M Turkish tweets, as well as 4.7M Bitcoin-related, 31M cryptocurrency, and 60M stock-related StockTwits posts (all StockTwits posts are in English), collected over the period 2007--2023.\footnote{Detailed statistics are provided in Appendix~\ref{app:data_app}.}

\subsection{Sentiment Labels}
StockTwits messages include native bullish/bearish labels, which we use as ground truth following prior work \cite{divernois2024stocktwits, cookson2020why}. Twitter lacks sentiment labels, so we infer them using GPT-5 \cite{openai2025gpt5}, validated against human annotations. Prior studies show that large language models achieve high-quality sentiment labeling \cite{he2023annollm, colavito2025benchmarking, gilardi2023chatgpt}. A subset of 2,700 tweets across the four languages was manually checked to verify label reliability. 
\section{Methods}

\subsection{Distributional Similarity}
We first examine how similarly financial communities use emojis by comparing the frequency distributions of the top 100 emojis across languages, platforms, and asset types. For each corpus pair, we compute four complementary divergence measures: Jensen--Shannon Distance (global information divergence), Total Variation (proportion of usage that differs), the Bhattacharyya Coefficient (distributional overlap), and Rank-Biased Overlap (agreement in the high-frequency head of the distribution). Together, these metrics capture global, proportional, overlapping, and rank-weighted similarities in emoji usage.

\subsection{Semantic Alignment}
To assess whether emojis preserve their contextual meanings across communities, we represent emoji semantics using multilingual contextual embeddings. For each corpus, all posts containing a given emoji are encoded with XLM-RoBERTa \cite{conneau2020xlmr}, and a centroid embedding is computed for each emoji. To compare semantic spaces across domains, we apply orthogonal Procrustes alignment and evaluate alignment using two metrics: (i) mean cosine similarity between corresponding emoji centroids and (ii) nearest-neighbor accuracy (NN@1 and NN@5), which measure how often an emoji’s closest counterpart in the other domain corresponds to the same symbol. Cosine similarity captures global semantic stability, while NN@$k$ reflects local neighborhood consistency.

\subsection{Sentiment Polarity Alignment}
We next evaluate whether emojis preserve their sentiment polarity across communities. For each emoji in each corpus, we estimate polarity as the proportion of positive posts among all posts containing that emoji. Polarity consistency across corpora is then assessed with three metrics: Spearman’s rank correlation (agreement in relative ordering), mean absolute polarity difference (magnitude of change), and flip rate (proportion of emojis whose polarity sign reverses). These metrics together quantify stability in sentiment polarity. 

\subsection{Cross-Community Transfer Experiments}
Finally, we test whether cross-community differences in emoji usage, meaning, or polarity affect downstream sentiment analysis. We perform zero-shot transfer experiments in which models are trained on one community and evaluated directly on another without retraining. We consider three transfer regimes: (i) cross-platform (Twitter vs.\ StockTwits), (ii) cross-asset (crypto vs.\ stocks), and (iii) cross-language (English, Spanish, Japanese, Turkish).

\subsubsection{Models}
We evaluate three model families that capture different levels of linguistic representation. A TF-IDF + Logistic Regression model provides an interpretable lexical baseline. XLM-RoBERTa serves as a strong multilingual contextual encoder suitable for short, emoji-rich posts. ByT5 \cite{xue-etal-2022-byt5} is included to test byte-level robustness to tokenization and vocabulary shifts across languages and platforms. We additionally evaluate a more recent encoder-based model, DeBERTa-v3-base fine-tuned for aspect-based sentiment analysis (ABSA), with results reported in Appendix~\ref{tab:gaps_cross_asset_deberta}--\ref{tab:gaps_cross_lingual_deberta}.

\subsubsection{Modalities}
To isolate the contribution of emojis, each model is trained under three input configurations: emoji-only (E), text-only (T), and combined text+emoji (TE). This yields a total of 27 experiments across domains, models, and modalities. For each setting, we report in-domain accuracy and the \emph{transfer gap}, defined as the drop in performance when evaluating on a target community. All corpora are balanced and split into separate training, validation, and test sets, and identical preprocessing is applied across domains.

Additional details are provided in the appendix, including corpus construction
and preprocessing (\ref{app:data_app}), sentiment labeling and validation
(\ref{app:sentiment_app}), detailed methods for emoji frequency, semantic,
polarity, and model transfer analyses
(\ref{app:frequency_app}--\ref{app:transfer_methods_app}),
statistical robustness procedures (\ref{app:statistical_app}),
and extended experimental results (\ref{app:transfer_app}).

\section{Results}

In this section, we present empirical results quantifying how financial communities differ in their emoji usage, meanings, and sentiment polarity, and whether such differences affect model transferability across communities. 

\subsection{Distributional Similarity} 
 
We first compare how frequently financial communities use emojis. Using the top 100 emojis per corpus, we compute Jensen--Shannon Distance (JSD), Total Variation (TV), Bhattacharyya Coefficient (BC), and Rank-Biased Overlap (RBO) (Table~\ref{tab:dist-sim}). These four metrics jointly provide complementary views: JSD and TV quantify global distributional differences, BC highlights shared usage despite frequency shifts, and RBO gives extra weight to agreement among the most frequent emojis---important given the heavy-tailed nature of emoji distributions. 

Across all comparisons, emoji frequency distributions differ substantially. Divergence is smallest for cross-asset pairs on the same platform, increases for cross-platform comparisons, and is largest for cross-language pairs. TV values indicate that roughly one-quarter to one-half of emoji usage proportions differ between domains. At the same time, high BC values and moderate RBO scores show that most emojis are shared across communities but used in different proportions. Overall, financial communication exhibits a partially shared emoji vocabulary, with frequency patterns shaped primarily by language and platform.
\begin{table}[t]
\centering
\caption{Emoji frequency distribution similarity across domains.
Higher JSD/TV indicates greater divergence; higher BC/RBO indicates greater similarity.}
\label{tab:dist-sim}
\small
\setlength{\tabcolsep}{4pt}
\begin{tabular}{lcccc}
\toprule
Comparison & JSD & TV & BC & RBO \\
\midrule
\textit{Cross-asset} \\
Stocks--Crypto (ST) & 0.28 & 0.26 & 0.95 & 0.69 \\
\midrule
\textit{Cross-platform} \\
ST--TW (BTC) & 0.32 & 0.28 & 0.92 & 0.64 \\
\midrule
\textit{Cross-language} \\
EN--ES (TW) & 0.42 & 0.38 & 0.87 & 0.45 \\
EN--JA (TW) & 0.51 & 0.50 & 0.80 & 0.25 \\
EN--TR (TW) & 0.41 & 0.38 & 0.88 & 0.45 \\
\bottomrule
\end{tabular}
\begin{flushleft}\footnotesize
ST = StockTwits, TW = Twitter, BTC = Bitcoin.
\end{flushleft}
\end{table}
\subsection{Semantic Alignment}

Frequency similarity does not reveal whether emojis convey similar meanings across communities. To assess semantic stability, we compare contextual emoji embeddings across domains using orthogonal Procrustes alignment. Mean cosine similarity measures the alignment of the overall contextual embedding spaces, while NN@1 and NN@5 quantify how consistently individual emojis align with their corresponding counterparts across corpora (Table~\ref{tab:emoji_alignment_main}). Intermediate values (NN@2--4) are reported in Table~\ref{tab:emoji_alignment_main_app}.

Although mean cosine similarity remains high in all settings, nearest-neighbor accuracy drops sharply in cross-platform and cross-lingual comparisons (particularly between English and Japanese). This pattern indicates that the global structure of emoji semantics is largely shared across communities, while the precise contextual semantics of specific emojis vary increasingly under platform and language shifts. 

\begin{table}[t]
\centering
\caption{Semantic alignment of emojis across domains.}
\label{tab:emoji_alignment_main}
\small
\setlength{\tabcolsep}{4pt}
\begin{tabular}{lcccc}
\toprule
Comparison & Mean Cosine & NN@1 & NN@5 \\
\midrule
\textit{Cross-asset} \\
Stocks--Crypto (ST) & 0.96 & 0.77 & 0.94 \\
\midrule
\textit{Cross-platform} \\
ST--TW (BTC) & 0.94 & 0.17 & 0.28 \\
\midrule
\textit{Cross-language} \\
EN--ES (TW) & 0.97 & 0.04 & 0.37 \\
EN--TR (TW) & 0.96 & 0.08 & 0.38 \\
EN--JA (TW) & 0.95 & 0.09 & 0.19 \\
\bottomrule
\end{tabular}
\begin{flushleft}\footnotesize
ST = StockTwits, TW = Twitter, BTC = Bitcoin.
\end{flushleft}
\end{table}

\subsection{Sentiment Polarity}

After assessing semantic similarity, we examine whether emojis preserve their sentiment polarity across communities. Semantic alignment does not guarantee affective consistency: the same emoji can express opposite sentiment even when its meaning is similar. For example, the \emoji{gem} emoji often signals resilience in financial discussions---positive when expressing confidence (“I will hold because I believe it will rise”) but negative when expressing stubbornness (“I will hold despite losses”). 

For each shared emoji, we compare its positive-to-negative usage ratios across corpora and evaluate polarity consistency using weighted Spearman correlation ($\rho$), weighted mean absolute polarity difference (MAUD$_w$), and weighted polarity flip rates. Frequency weighting reduces the influence of low-support emojis and yields more stable aggregate estimates.

As shown in Table~\ref{tab:emoji_alignment_all}, emoji sentiment rankings are strongly correlated across communities ($\rho_w=0.79$–$0.89$), indicating broadly consistent affective interpretation. Polarity shifts are modest (MAUD$_w\approx0.04$–$0.08$), and polarity flips are rare (below 8\%). Consistency is highest for cross-asset comparisons, decreases across platforms, and varies across languages. Overall, emoji sentiment polarity remains relatively stable across financial microblogging communities.

\begin{table}[t]
\centering
\caption{Emoji sentiment polarity consistency across domains.}
\label{tab:emoji_alignment_all}
\small
\setlength{\tabcolsep}{4pt}
\begin{tabular}{lccc}
\toprule
Comparison & $\rho_w$ & MAUD$_w$ & Flip$_w$ (\%) \\
\midrule
\textit{Cross-asset} \\
Stocks--Crypto (ST) & 0.89 & 0.044 & 2.5 \\
\midrule
\textit{Cross-platform} \\
ST--TW (BTC) & 0.81 & 0.086 & 7.9 \\
\midrule
\textit{Cross-language} \\
EN--ES (TW) & 0.73 & 0.081 & 7.1 \\
EN--JA (TW) & 0.85 & 0.064 & 3.5 \\
EN--TR (TW) & 0.79& 0.047 & 1.1 \\
\bottomrule
\end{tabular}
\begin{flushleft}\footnotesize
ST = StockTwits, TW = Twitter, BTC = Bitcoin.
\end{flushleft}
\end{table}

Additional visualizations of cross-community emoji usage and polarity are provided in Appendix Figures~\ref{fig:emoji_usage_app} and~\ref{fig:emoji_polarity_app}.

\subsection{Sentiment Model Transfer}

The preceding analyses quantify how emoji usage, semantics, and sentiment polarity vary across financial communities. We now examine whether these differences translate into performance gaps for sentiment analysis models. Specifically, we evaluate the zero-shot transferability of financial sentiment models trained on one community and tested on another across assets, platforms, and languages. 

We compare three model families: TF-IDF with logistic regression as an interpretable lexical baseline, XLM-R as a multilingual contextual encoder, and ByT5 as a byte-level transformer. Each model is evaluated under three input configurations: emoji-only (E), text-only (T), and combined text+emoji (TE), allowing us to isolate the contribution of emoji information to cross-domain generalization.

For each transfer setting, we report the transfer gap, defined as the difference between in-domain and out-of-domain accuracy. 

\subsubsection{Cross-asset}

Cross-asset transfer evaluates how well sentiment models trained on one asset community (cryptocurrencies) generalize to another (stocks) within the same platform and language. As shown in Table~\ref{tab:gaps_cross_asset_ci}, transferring across asset domains consistently reduces accuracy across all model families and input modalities, with transfer gaps ranging from approximately 2 to 11 percentage points. 

Text-only models exhibit the largest performance degradation. In contrast, emoji-only models transfer more smoothly, with substantially smaller accuracy gaps (all below 5\%), indicating that emojis encode sentiment signals that generalize better across asset types. Combining text and emoji yields the highest absolute accuracy and consistently reduces the transfer gap relative to text-only models, showing that emoji information complements textual features. 

Across model families, XLM-R shows the smallest transfer gaps overall, particularly in emoji-only and text+emoji settings, indicating that multilingual contextual representations capture relatively stable sentiment associations that transfer well across financial assets. 

\begin{table}[t]
\centering
\caption{Cross-asset transfer gaps in accuracy between StockTwits--Crypto (in-domain) and StockTwits--Stocks (out-of-domain).}
\label{tab:gaps_cross_asset_ci}
\small
\setlength{\tabcolsep}{4pt}
\begin{tabular}{llcc}
\toprule
Modality & Model & In-domain & $\Delta{\to}$ ST--Stocks \\
\midrule
Emoji      & ByT5    & 0.749 & 0.045 \\
           & XLM-R   & 0.718 & 0.020 \\
           & TF-IDF & 0.768 & 0.053 \\
\midrule
Text       & ByT5    & 0.783 & 0.092 \\
           & XLM-R   & 0.739 & 0.035 \\
           & TF-IDF & 0.845 & 0.106 \\
\midrule
Text+Emoji & ByT5    & 0.833 & 0.070 \\
           & XLM-R   & 0.791 & 0.034 \\
           & TF-IDF & 0.852 & 0.087 \\
\bottomrule
\end{tabular}
\begin{flushleft}\footnotesize
ST = StockTwits.
\end{flushleft}
\end{table}

\subsubsection{Cross-platform}

Cross-platform transfer evaluates how well sentiment models trained on one platform (StockTwits) generalize to another (Twitter) for the same asset and language. As shown in Table~\ref{tab:gaps_cross_platform_ci}, transferring across platforms leads to substantially larger accuracy drops than cross-asset transfer, with losses ranging from approximately 3 to 21 percentage points depending on the model and input modality.

Text-only models again exhibit the largest degradation. In contrast, emoji-only models transfer more effectively across platforms, with consistently smaller accuracy gaps. Among these, XLM-R shows near-zero transfer loss in the emoji-only setting, indicating that emoji--sentiment associations learned by multilingual contextual embeddings remain stable across platforms. ByT5 exhibits moderate transfer loss, while achieving higher in-domain accuracy. It is worth noting that \texttt{XLM-R} is pretrained on a general corpus of web data, possibly containing relatively limited finance-related social media content, which may explain its lower absolute performance despite strong transfer stability. 

Combining text and emoji improves transferability for all model families, consistently reducing cross-platform gaps relative to text-only models. Overall, these results indicate that platform shifts introduce substantial challenges for textual sentiment models, while emoji-based representations provide more robust cross-platform generalization.

\begin{table}[t]
\centering
\caption{Cross-platform transfer gaps in accuracy between StockTwits--BTC (in-domain) and Twitter--BTC (out-of-domain).}
\label{tab:gaps_cross_platform_ci}
\small
\setlength{\tabcolsep}{4pt}
\begin{tabular}{llcc}
\toprule
Modality & Model & In-domain & $\Delta{\to}$ TW--BTC \\
\midrule
Emoji      & ByT5    & 0.749 & 0.059 \\
           & XLM-R   & 0.718 & 0.004 \\
           & TF-IDF & 0.738 & 0.035 \\
\midrule
Text       & ByT5    & 0.783 & 0.209 \\
           & XLM-R   & 0.739 & 0.035 \\
           & TF-IDF & 0.831 & 0.191 \\
\midrule
Text+Emoji & ByT5    & 0.833 & 0.147 \\
           & XLM-R   & 0.791 & 0.022 \\
           & TF-IDF & 0.836 & 0.131 \\
\bottomrule
\end{tabular}
\begin{flushleft}\footnotesize
ST = StockTwits, TW = Twitter, BTC = Bitcoin.
\end{flushleft}
\end{table}

\subsubsection{Cross-language}

Cross-language transfer evaluates how well sentiment models trained on English data generalize to non-English financial communities on the same platform and asset. As shown in Table~\ref{tab:gaps_cross_lingual_ci}, cross-language transfer results in the largest performance degradation among all transfer settings, with accuracy losses exceeding 25\% for some combinations of target language, model, and input modality.

Across all models, incorporating emojis improves cross-language transfer. Emoji-only models exhibit the smallest transfer gaps, indicating that emoji-based sentiment signals are comparatively language-independent. Text-only models, in contrast, suffer the largest degradation, showing that textual sentiment cues are highly language-specific and do not generalize well across languages. Models trained on combined text+emoji inputs provide the best overall trade-off: while their transfer gaps remain larger than emoji-only models due to higher in-domain accuracy baselines, they are consistently smaller than those of text-only models, resulting in the highest absolute accuracy after transfer. This indicates that emojis stabilize text-based sentiment representations under cross-lingual domain shift. 
Among model families, XLM-R shows the strongest transfer robustness, reflecting the benefits of multilingual contextual pretraining that aligns sentiment representations across languages. TF--IDF exhibits the largest gaps, consistent with its reliance on exact token overlap and memorizing language-specific words; nevertheless, it retains non-trivial performance by exploiting non-textual tokens such as numbers and punctuation. ByT5 achieves intermediate robustness: while its byte-level encoding enables language-agnostic input processing, the absence of multilingual semantic pretraining leads to larger transfer losses, particularly for languages with highly distinct alphabets. 
Among target languages, transfer to Spanish yields the smallest gaps in text-based settings, consistent with its closer linguistic proximity to English, while this advantage is substantially reduced in emoji-based modalities, where sentiment meanings are more language-independent.

\begin{figure}[t]
  \centering
  \begin{subfigure}{0.48\columnwidth}
    \centering
    \includegraphics[width=\linewidth]{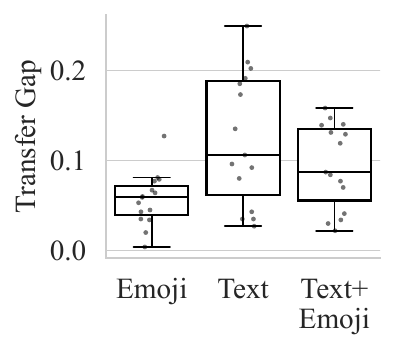}
    \caption{Modality}
    \label{fig:box-modality}
  \end{subfigure}

  \begin{subfigure}{0.48\columnwidth}
    \centering
    \includegraphics[width=\linewidth]{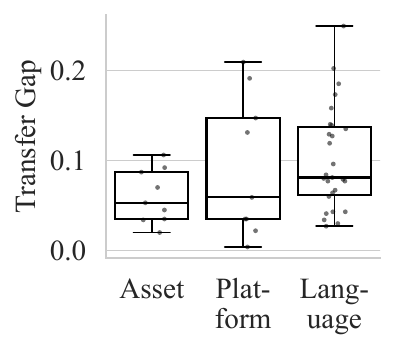}
    \caption{Regime}
    \label{fig:box-regime}
  \end{subfigure}\hfill
  \begin{subfigure}{0.48\columnwidth}
    \centering
    \includegraphics[width=\linewidth]{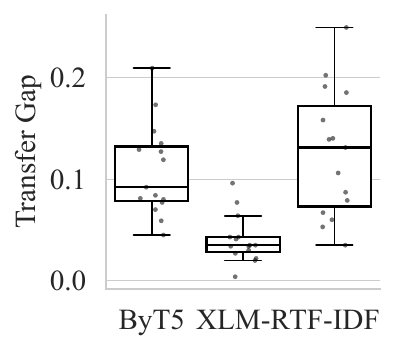}
    \caption{Model}
    \label{fig:box-model}
  \end{subfigure}

  \caption{Transfer gaps by modality, regime, and model.}
  \label{fig:box-all}
\end{figure}

\begin{table}[t]
\centering
\caption{Cross-language transfer gaps in accuracy between English (in-domain) and non-English corpora (out-of-domain).}
\label{tab:gaps_cross_lingual_ci}
\small
\setlength{\tabcolsep}{2.5pt}
\begin{tabular}{llcccc}
\toprule
Modality & Model & EN & $\Delta{\to}$ES & $\Delta{\to}$JA & $\Delta{\to}$TR \\
\midrule
Emoji      & ByT5    & 0.80 & 0.08 & 0.13 & 0.08 \\
           & XLM-R   & 0.77 & 0.06 & 0.03 & 0.04 \\
           & TF--IDF & 0.79 & 0.08 & 0.06 & 0.07 \\
\midrule
Text       & ByT5    & 0.90 & 0.08 & 0.14 & 0.17 \\
           & XLM-R   & 0.87 & 0.03 & 0.04 & 0.10 \\
           & TF--IDF & 0.85 & 0.19 & 0.25 & 0.20 \\
\midrule
Text+Emoji & ByT5    & 0.92 & 0.08 & 0.12 & 0.13 \\
           & XLM-R   & 0.90 & 0.03 & 0.04 & 0.08 \\
           & TF--IDF & 0.88 & 0.14 & 0.16 & 0.14 \\
\bottomrule
\end{tabular}
\end{table}

\section{Discussion}
\subsection{Findings and Implications}
Our results connect three layers---frequency distributions, semantic meaning, and sentiment polarity---to downstream model transfer performance. Emoji frequencies differ across communities, showing the largest divergence across languages and the smallest across assets. Despite these frequency shifts, the global semantic space remains stable (mean cosine > 0.93), while emoji-level similarity (NN@k) degrades mainly across languages and platforms. Sentiment polarity is highly consistent, with strong rank correlations between polarity distributions, small absolute differences, and high agreement in polarity signs across emojis. Together, these findings indicate a universally stable emoji semantic and sentiment structure, with community-specific usage patterns layered on top.

These three layers relate to the patterns of transfer we observe across regimes. Cross-asset comparisons (within the same platform and language) show the highest semantic alignment, strongest polarity agreement, and smallest divergence in emoji frequency distributions. They also have the smallest model transfer gaps, as shown in Fig.~\ref{fig:box-all}. 

The contrast between models further illustrates how model design shapes generalizability. XLM-R achieves the smallest transfer gaps. With its multilingual pretraining and contextual representation, it is more robust to community-specific vocabulary differences. ByT5 performs competitively within-domain and shows moderate robustness across domains, benefiting from its tokenizer-free byte-level encoding, which enables it to process unseen tokens, including emojis. However, without explicit multilingual semantics, its ability to generalize meaning across languages is limited; for instance, it can process Japanese but has little understanding of its semantics because Japanese characters are encoded very differently at the byte level compared to English characters. TF–IDF, by contrast, relies on direct co-occurrence of tokens and labels and therefore degrades more under shifts in vocabulary. Overall, the hierarchy XLM-R $\gg$ ByT5 $>$ TF-IDF follows the progression from lexical matching to multilingual contextual understanding.

Perhaps most notably, emojis consistently help models generalize. When combined with text, they reduce transfer gaps and improve accuracy across all regimes and models. This is consistent with their stable polarity and semantics even under language shifts. Importantly, emoji signals complement rather than replace textual ones: text+emoji models outperform text-only models, while emoji-only models remain strong for applications where textual learning is challenging. Emojis contribute a parallel, language-independent channel of sentiment information that stabilizes the model’s predictions when textual channels diverge.

Therefore, including emojis significantly reduces the need to retrain models. The cross-platform setting is especially interesting, suggesting that training emoji-based sentiment models on StockTwits data, which are labeled at a much higher rate, and applying them to the much more sparsely labeled Twitter data is feasible with limited loss. Emoji features also offer a compact, interpretable, faster-to-process signal. This supports the development of fast, explainable sentiment models that transcend cultural boundaries.

Our results position the paper between two seemingly opposing bodies of work: studies documenting wide cross-community variation in emoji use and meaning, and claims that emojis serve as near-universal sentiment signals. We concur with evidence of heterogeneity---frequency profiles diverge across communities, and cross-lingual NN@k alignment is low---consistent with cultural and platform effects. Yet we qualify this view with a finance-specific regularity: despite frequency shifts, the global semantic structure and sentiment polarity of emojis remain broadly stable across financial communities. This combination challenges the common inference that heterogeneity necessarily undermines transfer. In our setting, emojis consistently improve zero-shot generalization despite a certain degree of heterogeneity, and text+emoji models outperform text-only across regimes. We also advance the debate by distinguishing general ``emoji universality'' from domain-specific transferability, and by showing that, in finance, cultural convergence supports cross-cultural model transfer.

\subsection{Future Research}
Future research can build on these findings in several directions. A first step is to extend coverage across platforms and languages. Platforms such as Discord, Telegram, and Weibo, or other regional platforms, host significant communities whose emoji norms may differ from those of Twitter and StockTwits. Beyond sentiment, future work should investigate the broader communicative roles of emojis---including sarcasm and social signaling. Understanding these dimensions would help distinguish affective from pragmatic emoji use across communities and improve NLP models. Finally, a more causal analysis is needed to disentangle the effects of emoji frequency, semantics, and sentiment polarity on the model's transfer gap. Controlled experiments could identify which mechanisms most limit generalization.

\section{Conclusion}

We ask whether emojis provide \emph{transferable} sentiment signals across financial communities---spanning languages, assets, and platforms---and how community-specific variation impacts zero-shot model transfer. This matters for building efficient, cross-cultural sentiment systems in finance, where labeled data are scarce. 

Across all analyses---usage frequencies, semantics, sentiment polarity, and model transfer---a consistent pattern emerges.
Emoji behavior in finance reflects a shared communicative structure with domain-specific nuance. Emoji-sentiment metrics diverge more across languages and platforms than across assets, suggesting that community, rather than topic, drives variation.
Semantic alignment shows that global embedding structure is preserved, yet local neighborhoods shift, particularly in cross-lingual and cross-platform settings. 
Polarity analysis confirms that while most emojis retain their sentiment orientation, a few context-dependent symbols invert meaning depending on the community.
Finally, model transfer results integrate these observations: emoji inputs yield more stable cross-domain generalization than textual inputs, indicating that emojis---though not perfectly universal---encode robust sentiment signals that bridge financial subcultures.

Taken together, our findings provide clear answers to the guiding research questions. We find that financial communities rely on largely overlapping sets of emojis, yet their relative frequencies vary substantially across communities. Despite these differences in usage, the underlying semantic structure of emojis remains broadly aligned across domains. Global meanings are stable while local co-occurrences shift in subtle ways. Sentiment polarity, too, shows significant stability, with only a handful of emojis changing sentiment direction across communities. Finally, models trained with emoji inputs demonstrate notably stronger cross-domain generalization than their text-only counterparts, confirming that emojis serve as transferable sentiment signals in online financial communication. 

Our findings have both practical and conceptual implications. For practical model deployment, they suggest that emoji-based sentiment features are robust for transfer learning and can enhance the generalizability of financial NLP systems across markets and languages, reducing the need for domain-specific retraining---particularly useful given the scarcity of labeled data in many financial languages or platforms. From a conceptual perspective, our results nuance the debate on emoji universality. While prior studies outside finance reach mixed conclusions about whether emoji meanings are universal or culture-specific, none have examined financial communication---a domain we argue is more culturally convergent. We show that emoji representation in finance reflects both universal and community-specific patterns. Our findings also suggest that conclusions about transferability drawn from general social media may not extend to specialized contexts. This highlights the need for domain-sensitive analyses when assessing the universality of emojis.

\noindent\textbf{Data and Code Availability.}
Data and code required to reproduce the main results are available on Zenodo \citep{mahrous2026replication}.

\section{Limitations}

Nevertheless, several limitations remain. The analysis focuses on Twitter and StockTwits, and on four high-volume languages (EN, ES, JA, TR), leaving other platforms and linguistic groups unexplored. Only three model families were evaluated, excluding larger instruction-tuned LLMs. The study assumes single-label sentiment per post, and while correlations between drift and transfer gaps are strong, causal effects remain untested. Moreover, functions of emojis beyond bullish/bearish sentiment—such as humor or sarcasm—are not studied in this paper.

These limitations, however, do not undermine the main conclusions. Twitter and StockTwits represent two dominant venues for online financial discussion. The chosen languages cover diverse writing systems and cultural contexts, providing a strong basis for cross-lingual generalization. Evaluating three distinct model families---from lexical to contextual to byte-level---captures a broad design spectrum relevant to sentiment model transfer. Moreover, the models evaluated here may be more efficient than instruction-tuned LLMs for social media sentiment labeling \cite{colavito2025benchmarking, di2024performance}. This efficiency reinforces the practicality of emoji-based sentiment models for real-time financial applications. Finally, while causal analysis and non-sentiment uses of emojis remain an open research direction, they do not affect the observed stability of emoji meaning and polarity, which is consistent across datasets, languages, and model types.

\section{Ethical Considerations} \label{sec:ethics}

\subsection{Potential Risks}

This study poses minimal risk to individuals or communities. It uses only publicly available posts from Twitter and StockTwits and analyzes aggregate patterns of emoji use and sentiment rather than individual behavior. We do not infer personal attributes or make individual-level predictions. Although sentiment analysis tools could be misused for large-scale monitoring, this work is intended solely for academic research and does not support user-level profiling.

\subsection{Data Content and Privacy}

All data come from publicly accessible social media posts and exclude private or direct messages. During preprocessing, usernames, URLs, mentions, and hashtags were removed or normalized, retaining only linguistic content relevant to sentiment analysis, including emojis. All analyses are conducted at the aggregate level. No user identifiers are stored, analyzed, or released, and the shared artifacts do not permit re-identification.

The data may contain profanity or emotionally charged language, as is common in social media. Such content is neither highlighted nor analyzed at the individual level and does not affect the interpretation of our aggregate binary sentiment results.

In line with the GDPR~\cite{GDPR}, we process only publicly available data for academic research, following the principle of data minimization by retaining only material necessary for the study. Under institutional guidelines, this work qualifies as exempt secondary analysis of public data.

\section*{Acknowledgments}
This research was supported by award number 5940 from King Abdullah University of Science and Technology (KAUST).

\bibliography{refs} 

\appendix

\section{Data Construction and Preprocessing Details}
\label{app:data_app}

\begin{table*}[!t]
\centering
\resizebox{\textwidth}{!}{
\begin{tabular}{l r l l l r l l l}
\toprule
Dataset & Rows & Span & Prevalence & Intensity & Vocab & Eff.~N & Top-20 & Top-5 Emojis \\
\midrule
\multicolumn{9}{l}{\textbf{Twitter (Top-5 languages)}} \\
English    & 21{,}098{,}497 & 2007--2023 & 6.6\% [3.6, 10.6] & 2.26 [2.09, 2.38] & 1{,}568 & 161 & 47.9\% & {\centering \emoji{money-bag} \emoji{fire} \emoji{rocket} \emoji{check-mark-button} \emoji{right-arrow}} \\
Japanese   &   649{,}808 & 2010--2022 & 11.1\% [9.5, 12.4] & 2.25 [2.18, 2.32] &   579 & 139 & 46.9\% & {\centering \emoji{notes} \emoji{sparkles} \emoji{fire} \emoji{star} \emoji{thinking-face}} \\
Spanish    &   250{,}236 & 2011--2022 & 12.4\% [10.6, 13.7] & 2.65 [2.58, 2.72] &   461 & 132 & 46.9\% & {\centering \emoji{down-arrow} \emoji{up-arrow} \emoji{fire} \emoji{money-bag} \emoji{right-arrow}} \\
Turkish    &   225{,}426 & 2011--2022 & 12.6\% [11.1, 13.8] & 2.09 [2.04, 2.14] &   347 & 105 & 53.2\% & {\centering \emoji{face-with-tears-of-joy} \emoji{thumbs-up} \emoji{rocket} \emoji{check-mark-button} \emoji{fire}} \\
\midrule
\multicolumn{9}{l}{\textbf{StockTwits}} \\
Bitcoin  & 4{,}683{,}666  & 2017--2022 & 13.1\% [12.7, 13.4] & 2.59 [2.54, 2.64] & 1{,}162 & 108 & 53.8\% & {\centering \emoji{rocket} \emoji{fire} \emoji{money-bag} \emoji{chart-increasing} \emoji{chart-decreasing}} \\
Crypto   & 31{,}132{,}082 & 2012--2022 & 13.2\% [12.6, 13.8] & 2.94 [2.82, 3.07] & 1{,}666 & 96 & 56.4\% & {\centering \emoji{rocket} \emoji{money-bag} \emoji{fire} \emoji{gem-stone} \emoji{chart-increasing}} \\
Stocks   & 59{,}774{,}132 & 2009--2020 & 4.5\% [4.1, 4.9] & 2.44 [2.41, 2.47] & 1{,}305 & 111 & 54.0\% & {\centering \emoji{chart-increasing} \emoji{money-bag} \emoji{rocket} \emoji{fire} \emoji{chart-decreasing}} \\
\bottomrule
\end{tabular}
}
\caption{Descriptive statistics of the analyzed corpora. Each row summarizes a financial microblog dataset. \textit{Prevalence} = percentage of posts containing at least one emoji. \textit{Intensity} = average number of emojis per emoji-containing post. \textit{Vocab} = number of distinct emojis observed. \textit{Eff.~N} = $1 / \sum_i p_i^2$, where $p_i$ is the relative frequency of emoji $i$ (higher = more even usage). \textit{Top-20} = percentage of total emoji usage accounted for by the 20 most frequent emojis. The five most common emojis per dataset are shown as examples.}
\label{tab:emoji_stats_span_app}
\end{table*}

\subsection{Data Sources}

For Twitter, we merge two publicly available datasets of financial tweets~\cite{penasmartinez2023kaggle, alaix14kaggle}.
For StockTwits, we rely on two datasets: one containing stock-related microblogs obtained from~\cite{divernois2024stocktwits}, and another containing cryptocurrency-specific messages collected by the authors.

\subsection{Pre-processing steps}

Before analysis, all text was normalized by lowercasing and replacing URLs, user mentions, and hashtags with standardized tokens, while preserving emojis and punctuation that may carry sentiment, in line with \cite{renault2020comparison}.

Emojis were extracted using Unicode-aware regular expressions that handle multi-codepoint graphemes (e.g., family and flag sequences).
Skin tone modifiers, zero-width joiners, and variation selectors were removed to merge visually identical variants.

Exact duplicate messages were removed, and near-duplicates were filtered out using a text similarity hashing approach.
This step substantially reduced repetitive content, typically promotional posts or spam reposts, without affecting organic discussions.

\subsection{Language Identification}

Language identification employed an ensemble of fastText \cite{joulin2017bag}, LangID \cite{lui-baldwin-2012-langid}, MediaPipe \cite{mediapipe2024language}, and langdetect \cite{fedelopez2023langdetect} models, and tweets were retained only when a majority agreed on the detected language.
Posts were then grouped to enable subsequent cross-domain analyses spanning platform (Twitter vs.~StockTwits), asset type (stocks vs.~crypto), and language (English, Spanish, Japanese, Turkish) dimensions.

\section{Sentiment Labeling}
\label{app:sentiment_app}

\begin{figure}[htbp]
  \centering
  \includegraphics[width=0.7\columnwidth]{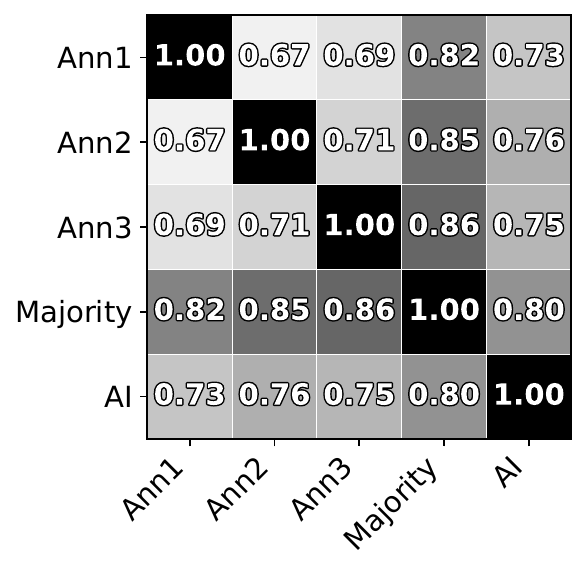}
  \caption{Pairwise agreement between human annotators, majority vote, and AI model. Values indicate the proportion of identical sentiment labels for each pair.}
  \label{fig:agreement-matrix_app}
\end{figure}

\subsection{StockTwits Native Labels}

For sentiment labeling, StockTwits provides self-labeled messages where posters can tag their posts as bullish (positive) or bearish (negative). These native labels were used directly as the ground-truth, following prior work showing that StockTwits self-reported sentiments are generally accurate and consistent with market activity \cite{divernois2024stocktwits, cookson2020why, oliveira2013predictability, audrino2020impact}. Throughout this study, sentiment is treated as a binary classification task, with either positive or negative labels.

\subsection{LLM-Based Sentiment Inference}

For Twitter, such labels are absent. Instead, we used GPT-5
\cite{openai2025gpt5} via the OpenAI API to infer tweet sentiment. Recent
studies have shown that large language models can approximate or exceed
human labeling quality in text classification tasks
\cite{he2023annollm,colavito2025benchmarking,gilardi2023chatgpt,nasution2024chatgpt}.
Our approach follows this emerging line of work, using GPT-5 to generate
initial sentiment labels for Twitter posts.

\subsection{Human Validation and Agreement}

The reliability of the AI-generated labels was evaluated through human
annotation. Three human annotators independently labeled a random subset
of 2{,}700 tweets spanning four languages (English, Spanish, Japanese,
and Turkish). For each language, the subset was balanced, with half of
the tweets labeled as positive and half as negative.

Annotators were instructed to assign a binary sentiment label based on the
expressed market outlook of each post. Tweets indicating optimism or expectation of price increase were labeled as positive,
while tweets expressing pessimism or expectation of price
decrease were labeled as negative. Annotators worked independently and
were blind to the AI-generated labels. Annotators were research collaborators and not crowdworkers; no recruitment
platform or payment scheme was involved.

Figure~\ref{fig:agreement-matrix_app} reports the pairwise agreement between
annotators, their majority vote, and the AI model. Agreement between the
GPT-based labels and the human majority reached 0.80, comparable to
human--human agreement (0.67--0.71). This consistency across all examined
languages indicates that GPT-based labeling achieves near-human
reliability and can be confidently used for large-scale multilingual
sentiment inference.

\section{Emoji Frequency and Distribution Metrics}
\label{app:frequency_app}

To quantify how similarly financial communities use emojis, we compared frequency distributions across corpora along three dimensions: platform (Twitter vs.~StockTwits), asset type (stocks vs.~crypto), and language (English, Spanish, Japanese, Turkish). For each corpus pair, we constructed the union of the top 100 most frequent emojis and computed four complementary metrics. Jensen--Shannon Distance (JSD) was used as an information-theoretic measure of global distributional divergence. Total Variation (TV) was used as a sanity check for JSD; TV also has a more interpretable meaning---indicating the proportion of emoji usage that differs between corpora. The Bhattacharyya Coefficient (BC) complements these by focusing on overlap rather than distance. This is useful when two corpora use many of the same emojis but in slightly different proportions---BC remains high while JSD may not. Rank-Biased Overlap (RBO) was used to address the heavy-tailed nature of emoji distributions. Unlike JSD/TV/BC, which treat all probabilities equally, RBO emphasizes agreement in the most frequent emojis and discounts the noisier long tail. This complementary combination avoids over-reliance on any single definition of similarity, offering information-theoretic, probabilistic, geometric, and rank-based similarity measures.

\section{Semantic Alignment Methodology}
\label{app:semantics_app}

For each corpus, we encoded all posts containing the top 100 most frequent emojis using the multilingual \texttt{XLM-RoBERTa-base} model~\cite{conneau2020xlmr}. This encoder was chosen for its strong cross-lingual sentence-level alignment \cite{conneau2020xlmr}, allowing us to represent emoji semantics from different languages in a shared multilingual vector space. For each emoji $e$, we randomly sampled $N$ posts (with $N{=}5000$ for platform/asset comparisons and $N{=}500$ for language comparisons) and computed the centroid vector---the mean of the normalized embeddings of those posts. Each centroid thus represents the average contextual meaning of that emoji in a given domain.

To ensure comparability, we filtered to emojis appearing at least $N$ times in both corpora. To compare the two embedding spaces, we applied \emph{orthogonal Procrustes alignment}, rotating one set of emoji centroids to best match the other in the least-squares sense while preserving pairwise distances. 

Semantic consistency was then quantified by three complementary metrics. \emph{Mean pairwise cosine similarity} measures the average alignment between corresponding emoji centroids after Procrustes transformation, reflecting overall semantic closeness between domains. \emph{Nearest-neighbor accuracy at $k{=}1$ (NN@1)} evaluates the proportion of emojis whose true counterpart in the other corpus is their closest semantic neighbor, while \emph{NN@5} relaxes this constraint to recognize cases where an emoji's meaning slightly shifts but remains semantically adjacent. 

Therefore, mean cosine similarity captures \textit{global} semantic alignment across corpora, while NN@$k$ highlights how well individual emojis preserve their \textit{local} pairwise meanings.

\section{Sentiment Polarity Analysis Methodology}
\label{app:polarity_app}

For each corpus, we aggregated all posts containing a given emoji and counted its positive and negative occurrences. The polarity of an emoji was then calculated as the proportion of positive occurrences. For each pair of corpora, we formed a comparison set by intersecting emojis that met minimum frequency thresholds on both sides---at least 300 total uses (with $\geq$30 positive and $\geq$30 negative) for platform or asset comparisons, and at least 120 total uses (with $\geq$12 positive and $\geq$12 negative) for language comparisons. 

To ensure that both positive and negative extremes were represented, we also included ``extreme-polarity tails'': the most positive and most negative emojis (top and bottom 100 for platform/asset pairs, and 50 for language pairs) that appeared in both corpora, even if they did not meet the support threshold. While such emojis may have limited evidence (e.g., only a few positive mentions), Jeffreys smoothing and frequency-based weighting were used to mitigate their noisy influence on aggregate metrics. 

For each pair of corpora, we computed the following metrics: Spearman's rank correlation to measure whether the relative ordering of emoji sentiment is consistent; Mean Absolute Unweighted Difference (MAUD) to quantify the average magnitude of polarity shift; and Flip Rate to assess the proportion of emojis whose polarity sign differs between corpora. An emoji was marked as a flip when the signs of its median polarity differed between corpora, and fewer than 5\% of bootstrap samples reversed this difference (i.e., the 95\% confidence interval of $\Delta\theta_e$ excluded zero). Weighted variants of the metrics were computed using harmonic-mean frequency weights, emphasizing emojis that were well represented in both corpora.

Annotators were instructed to label each post as positive (bullish) or negative (bearish) based on the expressed market sentiment.
Positive sentiment includes optimism, confidence, or expectation of price increase.
Negative sentiment includes pessimism, fear, or expectation of price decrease.
Neutral or ambiguous posts were resolved by choosing the dominant implied sentiment.
Annotators were blind to model predictions and labeled independently.

\section{Cross-Community Transfer Experiments Methodologies}
\label{app:transfer_methods_app}

The preceding analyses quantify differences in emoji usage, meaning, and polarity across domains. To evaluate whether these differences impact sentiment analysis in practice, we next perform cross-community transfer experiments, training models in one community and testing them in another without retraining. 

\subsection{Regimes: Cross-asset, Cross-platform, Cross-language}

We considered three regimes of domain shifts: \emph{cross-platform} (Twitter vs.~StockTwits), \emph{cross-asset} (crypto vs.~stocks), and \emph{cross-language} (English, Spanish, Japanese, Turkish). When studying one dimension, the other two were held constant to isolate its effect. For example, cross-platform analysis involved English posts on Bitcoin; cross-asset analysis involved posts on the same platform (StockTwits) in the same language (English); cross-language transfer involved posts on the same platform (Twitter) and on the same asset (Bitcoin). This design allows us to isolate the effect of each type of domain variation. 

\subsection{Models: TF–IDF, XLM-R, ByT5}

We used three models to capture different levels of linguistic representation. 

TF–IDF + Logistic Regression serves as a transparent, frequency-based baseline. It captures direct word and emoji co-occurrence patterns with sentiment, allows interpretable feature analysis, and provides a lower-bound reference for cross-domain transfer without pretrained contextual knowledge. 

XLM-RoBERTa, a multilingual transformer pretrained on large-scale web data (CC100) covering more than 100 languages, was selected as a strong contextual model for microblogs. Although its pretraining corpus does not specifically include Twitter data, its scale and diversity enable effective transfer to short and informal text, including emoji-rich posts. Moreover, its multilingual capability enables a fair comparison across language pairs by employing a shared subword vocabulary (SentencePiece) and a unified embedding space, where tokens from different languages are represented within the same vector space. This reduces the need for language-specific tokenization. The model therefore serves as an appropriate baseline for the task at hand \cite{conneau2020xlmr,barbieri2022xlmt}.

Finally, ByT5 was included to test transfer robustness at the byte level. Unlike subword-based models like RoBERTa, ByT5 directly encodes raw text as bytes, avoiding the dependence on tokenizer vocabularies that can bias cross-language transfer. ByT5 thus evaluates whether purely byte-level representations transfer more effectively than token-based representations. By not depending on a tokenizer, its transfer performance becomes more robust to non-standard or unseen tokens. Because all inputs are represented in the same byte space (0-255), different languages share a common low-level representation \cite{xue-etal-2022-byt5}.

All models were trained using standard configurations from prior work.
TF–IDF models used unigrams and bigrams with L2-regularized logistic regression (C=1.0).
XLM-R fine-tuning used a learning rate of 2e-5, batch size 32, AdamW optimizer, and 3 epochs.
ByT5 models were trained with a learning rate of 1e-4, batch size 16, and 3 epochs.
No hyperparameter search was conducted.

\subsection{Modalities: Emoji, Text, Text+Emoji}

To assess the role of emojis, each model was trained under three input configurations: E (emoji-only), T (text-only), and TE (text+emoji). These settings help us understand whether emoji information complements or substitutes textual sentiment cues under domain shift. In total, the setup yields 27 distinct experiments (3 domains × 3 models × 3 input configurations). For the cross-language transfer, each source language was evaluated against three distinct target languages. In addition to providing new insights, the variety of experiments also serves as a robustness check, showing that findings are consistent across different conditions. Each in-domain dataset contained 55,000 training, 5,000 validation, and 5,000 in-domain test samples. Transfer tests used 5,000 samples from the target domain. All datasets were balanced to have 50\% positive and 50\% negative samples. For each experiment, we report in-domain performance and the transfer gap, defined as the difference between in-domain and out-of-domain performance.

Each experiment followed a train-on-source, test-on-target setup for binary sentiment classification (positive vs. negative). Source data were split 80/20 into stratified train and validation sets, and identical preprocessing was applied across all domains. By stratifying, we ensure that both the training and validation sets maintain the same proportion of positive and negative samples as the full dataset. The dataset was balanced by undersampling the majority (positive) class. This was done because the original data was highly imbalanced, with only around 20\% of posts belonging to the negative class. Models were trained only on the source corpus and evaluated directly on the target corpus without retraining, allowing us to measure cross-domain transferability.

\section{Statistical Robustness Procedures}
\label{app:statistical_app}

\subsection{Bootstrapped Confidence Intervals}

Across analyses, 95\% confidence intervals (CIs) were estimated using non-parametric bootstrapping with 1,000 resamples. In each case, the unit of resampling matched the type of analysis. For descriptive statistics in Table~\ref{tab:emoji_stats_span_app}, monthly block bootstrapping was used, where entire month-level aggregates were resampled with replacement. For distributional distance and semantic similarity analyses, the shared emoji set between corpora was resampled with replacement to measure uncertainty due to which emojis are present; if one or two emojis dominate, their removal or duplication can shift the metric. For sentiment alignment, we used two bootstrap procedures, depending on the target uncertainty. For \emph{global} alignment metrics (Spearman~$\rho$, mean absolute polarity difference), we resampled the set of shared emoji types with replacement and recomputed the metric to obtain 95\% CIs. For \emph{per-emoji} polarity differences, we held an emoji $e$ fixed and resampled its posts with replacement within each corpus, recomputed its polarity, and formed a 95\% CI for $\Delta\theta_e$ (corpus~A minus corpus~B). An emoji was flagged as a robust flip when this interval excluded zero. For model transfer experiments, test samples were resampled (stratified by class to maintain label composition), and bootstrapped CIs for accuracy and related metrics were computed from the resulting distributions. 

Although temporal effects are not the main focus of this study, we repeated all experiments on time-aligned corpora (restricted to overlapping quarters). We also applied block bootstrapping at the month level, and all sampling was stratified by quarter to maintain temporal balance. Results remained unchanged within 95\% confidence intervals across all domains, indicating that temporal shifts do not materially affect our conclusions.

\subsection{Permutation Tests}
To assess whether observed differences were statistically meaningful, we used non-parametric permutation significance tests tailored to each analysis. 

For \emph{semantic alignment}, the test evaluated whether the observed mean cosine similarity between corpora was higher than expected if their embedding spaces were unrelated. In each permutation, emoji centroids from corpora \(A\) and \(B\) were randomly shuffled before computing the mean cosine similarity. The observed value was considered significant if it exceeded 95\% of the shuffled scores across 1,000 permutations. 

For \emph{sentiment alignment}, we tested whether the relationship in emoji polarities between corpora could arise by chance using a similar approach. Sentiment samples were randomly reassigned between corpora \(A\) and \(B\) for each emoji, and metrics such as weighted Spearman correlation, MAUD, and flip rate were recomputed for each permutation. 

Finally, for \emph{model transfer}, we tested whether the observed transfer gap---the difference between in-domain and cross-domain performance---could result from random variation. Domain assignments (source vs.~target) of test predictions were permuted 1,000 times, and accuracy gaps were recalculated to form a null distribution. The empirical \(p\)-value was obtained as the proportion of permuted gaps whose magnitude equaled or exceeded the observed gap.

\section{Extended Results}
\label{app:transfer_app}

\begin{figure*}[htbp]
    \centering
    \includegraphics[width=1.2\columnwidth]{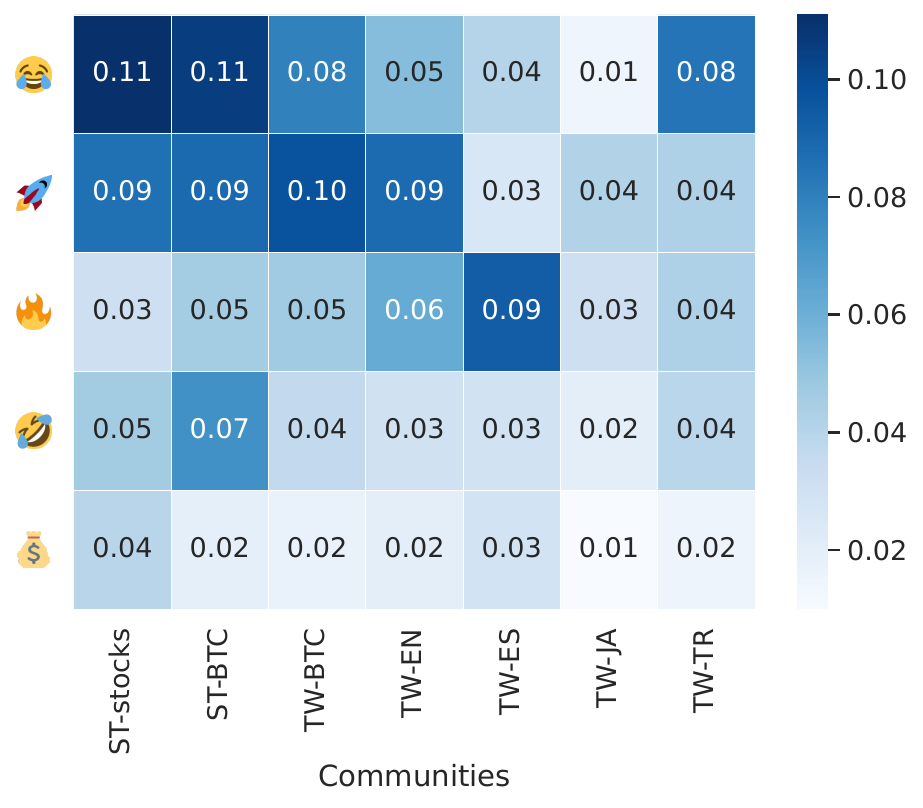}
    \caption{Emoji usage across communities (normalized share). Values are normalized within each community to allow fair comparison. Communities are StockTwits (stocks, BTC) and Twitter communities (BTC, EN, ES, JA, TR).}
    \label{fig:emoji_usage_app}
\end{figure*}

\begin{figure*}[htbp]
    \centering
    \includegraphics[width=1.2\columnwidth]{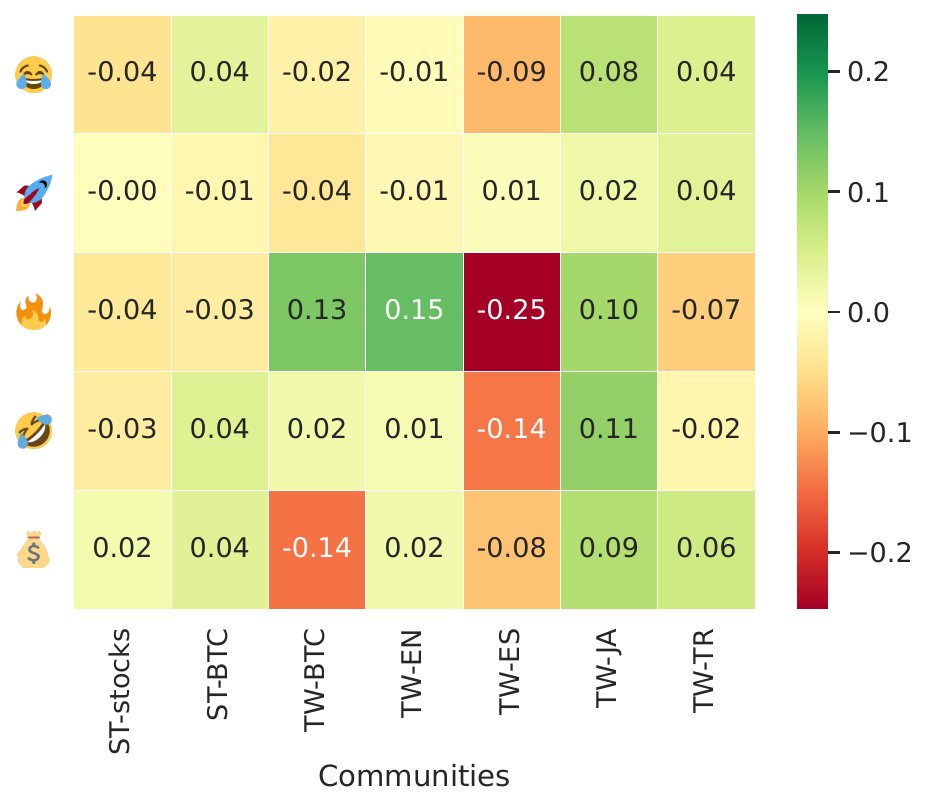}
    \caption{Emoji polarity across communities (centered scores). Values are centered relative to the global mean polarity of each emoji, highlighting cross-community semantic shifts.}
    \label{fig:emoji_polarity_app}
\end{figure*}

This section provides complementary or extended results.
Figures~\ref{fig:emoji_usage_app} and~\ref{fig:emoji_polarity_app} provide complementary visualizations of emoji variation across communities, showing normalized usage shares and centered polarity scores for five commonly used emojis with significant usage variation.

Tables~\ref{tab:dist-sim-five_app}--\ref{tab:gaps_cross_lingual_ci} report extended versions of the corresponding main-text tables, adding information about vocabulary sizes, 95\% bootstrap confidence intervals, permutation-based significance tests, and additional breakdowns omitted for space. 

\begin{table}[htbp]
\centering
\caption{Cross-asset transfer gaps in accuracy for DeBERTa-v3 (ABSA) between StockTwits--Crypto (in-domain) and StockTwits--Stocks (out-of-domain).}
\label{tab:gaps_cross_asset_deberta}
\small
\setlength{\tabcolsep}{4pt}
\begin{tabular}{lcc}
\toprule
Modality & Baseline (ST--Crypto) & $\Delta{\to}$ ST--Stocks \\
\midrule
Emoji      & 0.754 & 0.050 \\
Text       & 0.812 & 0.055 \\
Text+Emoji & 0.858 & 0.059 \\
\bottomrule
\end{tabular}
\end{table}

\begin{table}[htbp]
\centering
\caption{Cross-platform transfer gaps in accuracy for DeBERTa-v3 (ABSA) between StockTwits--BTC (in-domain) and Twitter--BTC (out-of-domain).}
\label{tab:gaps_cross_platform_deberta}
\small
\setlength{\tabcolsep}{4pt}
\begin{tabular}{lcc}
\toprule
Modality & In-domain & $\Delta{\to}$ TW--BTC \\
\midrule
Emoji      & 0.726 & 0.021 \\
Text       & 0.802 & 0.045 \\
Text+Emoji & 0.847 & 0.056 \\
\bottomrule
\end{tabular}
\end{table}

\begin{table}[htbp]
\centering
\caption{Cross-language transfer gaps in accuracy for DeBERTa-v3 (ABSA) between English (in-domain) and non-English corpora (out-of-domain).}
\label{tab:gaps_cross_lingual_deberta}
\small
\setlength{\tabcolsep}{3pt}
\begin{tabular}{lcccc}
\toprule
Modality & EN & $\Delta{\to}$ES & $\Delta{\to}$JA & $\Delta{\to}$TR \\
\midrule
Emoji      & 0.770 & 0.064 & 0.048 & 0.048 \\
Text       & 0.911 & 0.065 & 0.085 & 0.316 \\
Text+Emoji & 0.931 & 0.047 & 0.072 & 0.205 \\
\bottomrule
\end{tabular}
\end{table}
\begin{table*}[!htbp]
\centering
\caption{Emoji frequency distribution similarity. 
$V$ denotes the union of the Top--100 emojis from each corpus, and $|V|$ its size (unique emojis). 
Values show observed scores with 95\% bootstrap CIs; * $p<.05$ (permutation). 
JSD = Jensen–Shannon distance, TV = Total Variation, BC = Bhattacharyya coefficient, RBO = Rank-Biased Overlap.}
\label{tab:dist-sim-five_app}
\small
\begin{tabular}{lcccccc}
\toprule
Pair & $|V|$ & JSD & TV & BC & RBO \\
\midrule
ST–Crypto & 121 & 0.275* [.274, .277] & 0.256* [.253, .257] & 0.945* [.944, .945] & 0.689* [.685, .693] \\
BTC ST–TW & 131 & 0.318* [.316, .320] & 0.284* [.282, .286] & 0.923* [.922, .924] & 0.635* [.634, .640] \\
EN–ES TW  & 138 & 0.417* [.413, .423] & 0.375* [.368, .380] & 0.867* [.864, .870] & 0.447* [.446, .470] \\
EN–JA TW  & 140 & 0.512* [.510, .513] & 0.502* [.499, .503] & 0.803* [.802, .804] & 0.254* [.232, .254] \\
EN–TR TW  & 129 & 0.405* [.402, .411] & 0.382* [.380, .390] & 0.877* [.873, .879] & 0.451* [.392, .463] \\
\bottomrule
\end{tabular}
\begin{flushleft}\footnotesize
ST = StockTwits, TW = Twitter. Larger JSD/TV = greater divergence; larger BC/RBO = greater similarity.
\end{flushleft}
\end{table*}
\begin{table*}[!htbp]
\centering
\caption{Semantic alignment of shared emojis across platforms and languages. $N$ is the number of emojis shared between corpora. 
* $p{<}.01$ (permutation). 
NN@$k$ measures the fraction of emojis whose correct counterpart appears among the top-$k$ nearest neighbors after alignment.}
\label{tab:emoji_alignment_main_app}
\small
\setlength{\tabcolsep}{3.5pt}
\begin{tabular}{l r c ccccc}
\toprule
Comparison & $N$ & Mean Cosine & NN@1 & NN@2 & NN@3 & NN@4 & NN@5 \\
\midrule

\textit{Cross-asset} \\
Stocks--Crypto (ST)
& 48 & 0.955* 
& 0.77* & 0.89* & 0.89* & 0.91* & 0.94* \\

\midrule
\textit{Cross-platform} \\
ST--TW (BTC) 
& 46 & 0.935* 
& 0.17* & 0.20* & 0.24* & 0.24* & 0.28* \\

\midrule
\textit{Cross-language} \\
EN--ES (TW)
& 27 & 0.965* 
& 0.04* & 0.20* & 0.29* & 0.29* & 0.37* \\
EN--TR (TW)
& 24 & 0.959* 
& 0.08* & 0.20* & 0.25* & 0.38* & 0.38* \\
EN--JA (TW)
& 47 & 0.954* 
& 0.09* & 0.14* & 0.17* & 0.18* & 0.19* \\

\bottomrule
\end{tabular}
\end{table*}

\begin{table*}[!htbp]
\centering
\caption{Cross-asset transfer gaps in accuracy between StockTwits--Crypto (in-community) and StockTwits--Stocks (out-of-community). 
$\Delta$ denotes the transfer gap, which is the drop in accuracy when transferring a model trained on one asset community to another. 
Baseline accuracies are shown alongside transfer gaps with 95\% confidence intervals.}
\label{tab:gaps_cross_asset_ci_app_app}
\small
\setlength{\tabcolsep}{3.5pt}
\begin{tabular}{llcc}
\toprule
Modality & Model & Baseline (ST--Crypto) & $\Delta{\to}$ST--Stocks [95\% CI] \\
\midrule
Emoji      & ByT5    & 0.749 & 0.045* [0.030, 0.061] \\
            & XLM-R   & 0.718 & 0.020* [0.004, 0.037] \\
            & TF--IDF & 0.768 & 0.053* [0.048, 0.058] \\
\midrule
Text       & ByT5    & 0.783 & 0.092* [0.076, 0.107] \\
            & XLM-R   & 0.739 & 0.035* [0.019, 0.051] \\
            & TF--IDF & 0.845 & 0.106* [0.102, 0.110] \\
\midrule
Text+Emoji & ByT5    & 0.833 & 0.070* [0.057, 0.083] \\
            & XLM-R   & 0.791 & 0.034* [0.020, 0.048] \\
            & TF--IDF & 0.852 & 0.087* [0.083, 0.091] \\
\bottomrule
\end{tabular}
\end{table*}

\begin{table*}[!htbp]
\centering
\caption{Cross-platform transfer gaps in accuracy between StockTwits--BTC (in-community) and Twitter--BTC (out-of-community). 
$\Delta$ denotes the transfer gap, which is the drop in accuracy when transferring a model trained on one platform to another for the same asset. 
Baseline accuracies are shown alongside transfer gaps with 95\% confidence intervals.}
\label{tab:gaps_cross_platform_ci_app}
\small
\setlength{\tabcolsep}{3.5pt}
\begin{tabular}{llcc}
\toprule
Modality & Model & Baseline (ST--BTC) & $\Delta{\to}$TW--BTC [95\% CI] \\
\midrule
Emoji & ByT5 & 0.749 & 0.059* [0.042, 0.077] \\
            & XLM-R   & 0.718 & 0.004 [−0.021, 0.029] \\
            & TF--IDF & 0.738 & 0.035* [0.030, 0.040] \\
\midrule
Text       & ByT5    & 0.783 & 0.209* [0.184, 0.234] \\
            & XLM-R   & 0.739 & 0.035* [0.010, 0.059] \\
            & TF--IDF & 0.831 & 0.191* [0.187, 0.196] \\
\midrule
Text+Emoji & ByT5    & 0.833 & 0.147* [0.123, 0.170] \\
            & XLM-R   & 0.791 & 0.022 [−0.001, 0.045] \\
            & TF--IDF & 0.836 & 0.131* [0.126, 0.135] \\
\bottomrule
\end{tabular}
\end{table*}

\begin{table*}[!htbp]
\centering
\caption{Cross-language transfer gaps in accuracy between English (in-community) and non-English corpora (out-of-community). 
$\Delta$ denotes the transfer gap, which is the drop in accuracy when transferring a model trained on English messages to another language. 
Baseline accuracies are shown alongside transfer gaps with 95\% confidence intervals.}
\label{tab:gaps_cross_lingual_ci_app}
\small
\setlength{\tabcolsep}{3.5pt}
\begin{tabular}{llcccc}
\toprule
Modality & Model & Baseline (EN) & $\Delta{\to}$ES [95\% CI] & $\Delta{\to}$JA [95\% CI] & $\Delta{\to}$TR [95\% CI] \\
\midrule
Emoji      & ByT5    & 0.803 & 0.081* [0.073, 0.088] & 0.127* [0.119, 0.135] & 0.077* [0.069, 0.084] \\
            & XLM-R   & 0.767 & 0.064* [0.056, 0.073] & 0.034* [0.026, 0.042] & 0.043* [0.035, 0.051] \\
            & TF--IDF & 0.794 & 0.079* [0.066, 0.091] & 0.060* [0.048, 0.072] & 0.067* [0.054, 0.080] \\
\midrule
Text       & ByT5    & 0.899 & 0.080* [0.076, 0.085] & 0.135* [0.130, 0.141] & 0.173* [0.167, 0.179] \\
            & XLM-R   & 0.867 & 0.027* [0.023, 0.032] & 0.043* [0.038, 0.049] & 0.096* [0.090, 0.102] \\
            & TF--IDF & 0.853 & 0.185* [0.171, 0.197] & 0.249* [0.236, 0.262] & 0.202* [0.189, 0.216] \\
\midrule
Text+Emoji & ByT5    & 0.922 & 0.084* [0.080, 0.088] & 0.119* [0.115, 0.123] & 0.129* [0.124, 0.133] \\
            & XLM-R   & 0.896 & 0.030* [0.026, 0.033] & 0.041* [0.037, 0.045] & 0.077* [0.073, 0.082] \\
            & TF--IDF & 0.877 & 0.140* [0.127, 0.152] & 0.158* [0.147, 0.171] & 0.139* [0.127, 0.151] \\
\bottomrule
\end{tabular}
\end{table*}

\end{document}